\documentclass[letterpaper, 10 pt, conference]{ieeeconf}  

\IEEEoverridecommandlockouts                              

\usepackage{graphicx}
\usepackage{stfloats}
\usepackage{subcaption}
\usepackage{amsmath}
\usepackage{amssymb}
\usepackage{cleveref}
\usepackage{multirow} 
\usepackage{float}
\usepackage{cite}
\usepackage{hhline}
\usepackage{booktabs}
\usepackage{threeparttable}

\title{\LARGE \bf
BronchoCopilot: Towards Autonomous Robotic Bronchoscopy via Multimodal Reinforcement Learning
}

\author{Jianbo Zhao$^{1,2}$, Hao Chen$^{2,1}$, Qingyao Tian$^{1}$, Jian Chen$^{1}$, Bingyu Yang$^{1}$ and Hongbin Liu$^{1,3,4}$%
\thanks{*This work was supported by the Institute of Automation, Chinese Academy of Sciences. (Corresponding author: Hongbin Liu)}%
\thanks{$^{1}$Institute of Automation, Chinese Academy of Sciences, Beijing 100190, China {\tt\small (\{zhaojianbo2022, chenhao2020, tianqingyao2021, chenjian2020, yangbingyu2022, liuhongbin\}@ia.ac.cn)}}%
\thanks{$^{2}$ School of Artificial Intelligence, University of Chinese Academy of Sciences, Beijing 100049, China}%
\thanks{$^{3}$ Centre of AI and Robotics (CAIR), Hong Kong Institute of Science \& Innovation, Chinese Academy of Sciences, Hongkong, China}%
\thanks{$^{4}$ School of Engineering and Imaging Sciences, King’s College London, UK}%
}

\begin{document}
\bibliographystyle{ieeetr}

\maketitle
\thispagestyle{empty}
\pagestyle{empty}

\begin{abstract}
 Bronchoscopy plays a significant role in the early diagnosis and treatment of lung diseases. This process demands physicians to maneuver the flexible endoscope for reaching distal lesions, particularly requiring substantial expertise when examining the airways of the upper lung lobe. With the development of artificial intelligence and robotics, reinforcement learning (RL) method has been applied to the manipulation of interventional surgical robots. However, unlike human physicians who utilize multimodal information, most of the current RL methods rely on a single modality, limiting their performance. In this paper, we propose BronchoCopilot, a multimodal RL agent designed to acquire manipulation skills for autonomous bronchoscopy. BronchoCopilot specifically integrates images from the bronchoscope camera and estimated robot poses, aiming for a higher success rate within challenging airway environment. We employ auxiliary reconstruction tasks to compress multimodal data and utilize attention mechanisms to achieve an efficient latent representation of this data, serving as input for the RL module. This framework adopts a stepwise training and fine-tuning approach to mitigate the challenges of training difficulty. Our evaluation in the realistic simulation environment reveals that BronchoCopilot, by effectively harnessing multimodal information, attains a success rate of approximately 90\% in fifth generation airways with consistent movements. Additionally, it demonstrates a robust capacity to adapt to diverse cases.

\end{abstract}

\section{INTRODUCTION}

Bronchoscopy has been instrumental in the inspection and diagnosis of lung diseases \cite{r1}, \cite{r2}. It is a surgical procedure that allows medical professionals to visually examine the lungs and airways. Physicians are required to manipulate flexible, non-linear surgical instruments carefully through the airways to reach distal lesions, implying a requirement for extensive experience and skills. Robotic bronchoscopy platform \cite{r4} has emerged to alleviate difficulties of sensing and control for physicians, enhancing the diagnostic rate while reducing operational risks, such as discomfort or bleeding \cite{r3}. Nevertheless, due to the demands for precision and safety, mastering the platform still necessitates high training costs. As a result, current platform is expected to have higher-level autonomy, performing more complex tasks with enhanced success rate and consistent motion \cite{r4}. 

   \begin{figure}[t]
      \centering
      \includegraphics[scale=0.365]{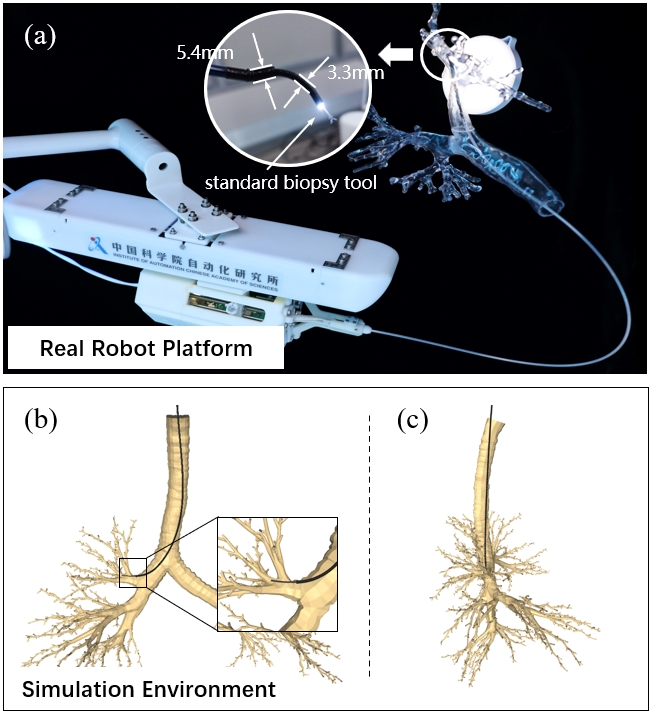}
      \caption{(a) Real surgical scenario: The operator is controlling the insertion of the robot bronchoscope. (b), (c) The simulation environment, includes the 3D airway model and the simulated dual-segment flexible endoscopic robot.}
      \label{figurelabel}
   \end{figure}

  Many navigation systems have been proposed for robot-assisted bronchoscopy, aiming to alleviate cognitive and physical stress on physicians. This allows them to redirect their focus from the manipulation of surgical instruments to making more advanced intervention decisions and diagnoses. Previous efforts include kinematics-based \cite{r5}, \cite{r6} and image-based \cite{r7}, \cite{r8} motion planners, which have significantly enhanced diagnostic rates. However, these methods require sophisticated manual designs, and they overlooked the interaction between the robot and the airway wall, but it is common because the robot needs support from the wall in tortuous airways. In recent years, propelled by advancements in artificial intelligence and robotics, reinforcement learning (RL) has been applied to surgical robots, enhancing the surgical autonomy \cite{r9, r10, r11, r12, r13, r14}. While showing benefits such as safety and transferability, current RL methods often struggle in complex tasks. Contrary to human physicians who leverage multimodal information for decision-making, RL methods typically concentrate on a single modality, either vision \cite{r9}, \cite{r10} or proprioception (including position and orientation) \cite{r11, r12, r13}. This challenge arises due to the heterogeneity and inconsistency of multimodal information, and the indirect reward feedback in RL training further amplifies its complexity \cite{r14}, \cite{r15}.

Inspired by the strategies employed by experienced surgeons, in this paper, we propose \textbf{BronchoCopilot}, a RL-based agent leveraging multimodal information for autonomous robotic bronchoscopy. Specifically, BronchoCopilot devises manipulation strategy to a target by bronchoscope camera images and estimated robot pose, with enhanced success rate, consistency and safety in complex airway environments. To address the challenges posed by the heterogeneity of different data modalities while maximizing their complementary nature, auxiliary reconstruction tasks \cite{r18} are introduced to obtain low-dimensional representations of multimodal data. Furthermore, we employ a cross-modal attention mechanism to dynamically adjust the significance of different modalities. Through a staged training regime and stepwise fine-tuning, we tackle the convergence challenges inherent in multimodal reinforcement learning algorithms.

As a proof of concept, the training and evaluation of BronchoCopilot are carried out in a realistic simulation environment, which is elaborated in the section III. A. The agent is designed for the dual-segment flexible endoscopic robot platform derived from \cite{r17}. As depicted in Fig. 1(a), the robot features an outer sheath and an inner endoscope, each actuated by three cables. Coordinated actions between these components are crucial for navigating to distal lesions, particularly within the finer airways of the upper lung lobe. The main contributions of this work are as follows:

\begin{itemize}

\item To the best of our knowledge, BronchoCopilot is the first work to use multimodal reinforcement learning for autonomous interventional surgical robot manipulation. 

\item We propose a novel algorithm framework that employs cross-attention and stepwise training, to fuse modalities with heterogeneity and alleviate the convergence difficulty.

\item Detailed experiments demonstrate that through our method, multimodal information can improve agent's performance in complex scenarios, as well as its generalization capability across various cases.

\end{itemize}

   \begin{figure}[thpb]
      \centering
      \vspace{5pt}
      \includegraphics[scale=0.38]{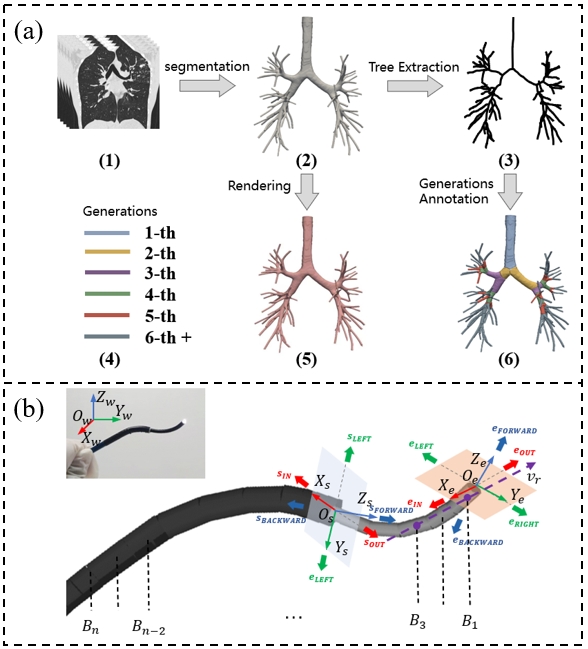}
      \caption{The establishment of the simulation environment. (a) The creation of the airway model, including segmentation from preoperative CT scans, bronchial tree extraction and rendering in the simulator, and we visualize the airway's generations based on tree branching. (b) The FEM modeling and the action space of the robot. The purple arrow $v_{r}$ denotes the orientation vector of the robot.}
      \label{figurelabel}
   \end{figure}

\section{RELATED WORK}

\subsection{RL Methods in Interventional Surgical Robots}

Interventional surgical robots are widely used in minimally invasive surgeries (MIS), typically equipped with flexible catheters or needles to navigate through vessels, cavities, or surgical openings \cite{r19}. They enable precise and tremor-free continuous operations. Nevertheless, due to the non-linearity of the instruments and the constraints of narrow working spaces, traditional controllers struggle to overcome the operational challenges posed by surgical robots \cite{r20}. Numerous efforts have applied RL methods to the manipulation of interventional surgical robots, include path planning \cite{r10}, \cite{r11}, \cite{r21}, surgical training systems \cite{r9}, \cite{r22}, and surrogate surgeon operations \cite{r12}, \cite{r13}. These methods typically rely on single-modal surgical information, unlike human physicians often requiring multimodal feedback to perform their manipulations. As a result, these methods are mostly confined to simpler tasks and scenarios, lacking the capability to learn in more complex surgical environments and instrument behaviors.

\subsection{Autonomous Robotic Bronchoscopy}

The integration of artificial intelligence and control theory with bronchoscopy significantly enhances the autonomy of robot-assisted bronchoscopy platforms, reducing training costs and workloads on physicians \cite{r23}. For example, the Alterovitz team \cite{r24}, \cite{r5} has made significant contributions to steerable needle lung robot, designing motion planners based on needle kinematics, with bronchoscope navigation being one stage in the overall process. However, these efforts did not specifically consider the deformation caused by bronchoscope contact with the airway wall, restricting the maneuverability of the bronchoscope. By leverage technologies of computer vision, researchers have developed a range of image-based guiding systems \cite{r7}, \cite{r8}, \cite{r25}. These systems typically utilize endoscopic images or fluoroscopy to estimate robot poses and devise insertion strategies, raising the precision and efficiency of the procedure. Nevertheless, these methods exhibit unsatisfactory performance in upper lung lobe interventions, which require pre-bending or retraction maneuvers \cite{r16}. Additionally, due to their complete reliance on visual information, limitations in perspective and estimation errors may raise concerns regarding surgical safety.

   \begin{figure*}[thpb]
      \centering
      \vspace{5pt}
      \includegraphics[width=0.98\linewidth]{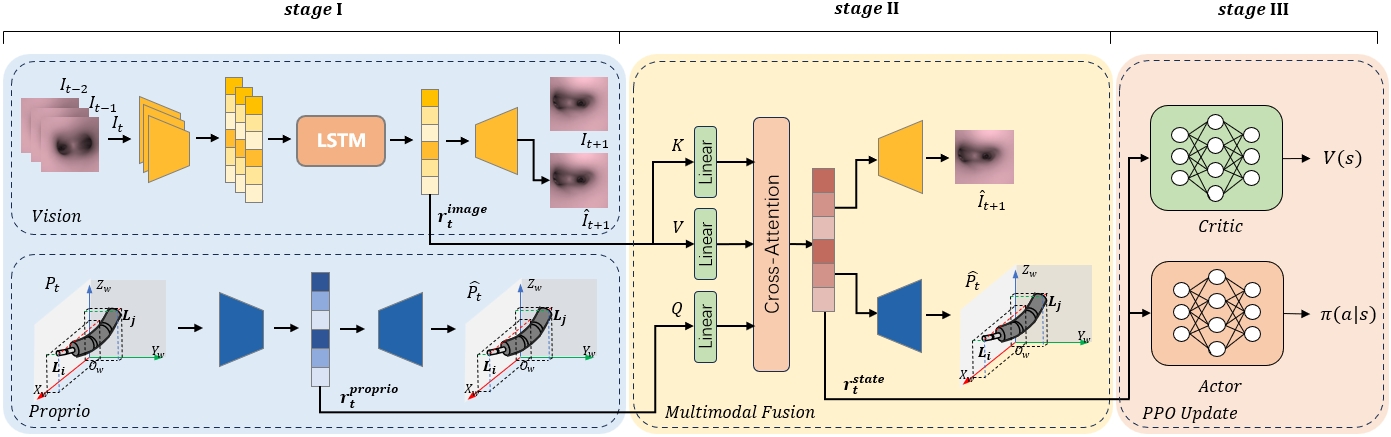}
      \caption{The Architecture of our method. The network takes data from three different modalities as input and outputs the manipulation policy. The entire architecture is trained in stages. In \textbf{\textit{stage} I}, it encodes multimodal information to low-dimensional embeddings though reconstruction tasks. In \textbf{\textit{stage} II}, it fuses multimodal embeddings into state representation as the input of \textbf{\textit{stage} III}, with the loss from subsequent tasks used to fine-tune the parameters of the front stage's network.}
      \label{figurelabel}
   \end{figure*}

\subsection{Multimodal Reinforcement Learning}

Multimodal models outperform single-modal ones, as confirmed by both theory and experiments \cite{r31}, \cite{r32}. The complementary of heterogeneous sensor modalities has been explored for training decision models \cite{r33}, \cite{r34}, \cite{r35}. The key to multimodal RL lies in how to obtain latent representations suitable for RL tasks \cite{r15}. It is easy when labeled data is available, but the indirect feedback in RL training makes it more challenging. We adapted the pre-training approach similar to \cite{r18}, leveraging auxiliary reconstruction tasks to obtain compressed and unified representations of multimodal information. These representations are further fine-tuned during the RL training phase. Furthermore, many works encode multimodal inputs using various encoders and then fuse them through summation or concatenation \cite{r36}, \cite{r37}. This approach may mask and introduce ambiguity in inter-modal information. In this paper, we explored a cross-attention based multimodal fusion approach \cite{r38}, \cite{r39} to integrate inter-modal information while addressing dynamic requirements.

\section{METHOD}

\subsection{Realistic Simulation Environment Design} 
Due to extended training durations and potential risky behaviors, direct training in real-world settings is costly. As illustrated in Fig. 1(b), we have developed a simulation environment tailored to provide detailed and authentic descriptions of actual bronchoscopy procedures. 

To construct the airway model, as shown in Fig. 2(a), we employ a segmentation approach \cite{r26} to extract lung anatomy from preoperative CT scans. The model is refined to generate surface mesh as collision environment, which is composed of multiple triangles. Subsequently, the obtained surface mesh is displayed by OpenGL \cite{r27}, where high-quality, realistic visual rendering is applied, including additional effects such as the reflection on the surface of the inner wall and the illumination from the front camera. We consider the bronchial model as rigid for the purpose of collision detection within the environment. The airway centerlines are extracted by VMTK \cite{r28} to serve as reference paths, then the airway generations can be determined. 

For the simulation of the robot, we use a finite element beam model to simulate the deformation of the robot and consider the effects of multiple contact points with the environment, similar to the SOFA \cite{r29}, \cite{r30}. In the simulation we use parameters obtained from real robot: The Young's modulus of the sheath is $510 \, \text{MPa}$, the length is $0.7 \, \text{m}$, the area is $9.30 \times 10^{-6} \, \text{m}^{2}$, and the moment of inertia is $19.233 \times 10^{-12} \, \text{m}^4$. Correspondingly, the corresponding parameters of the endoscope part are $307 \, \text{MPa}$, $0.7 \, \text{m}$, $2.83 \times 10^{-6} \, \text{m}^2$, $1.817 \times 10^{-12} \, \text{m}^4$.

It's essential that multimodal data is easy to access in the real environment. In this work, the visual information comprises endoscopic images obtained through a camera mounted at the robot's tip. The robot's proprioceptive information can be acquired by bronchoscopic localization systems like electromagnetic (EM). Furthermore, to capture the nonlinear kinematic characteristics of the robot, advanced shape perception technique \cite{r48} is employed for overall pose estimation of the robot.

\subsection{Problem Statement} 

The problem can be stated as follows: In the realistic simulation environment, a continuous flexible robot moves within the airways and reaches distal target. BronchoCopilot needs to determine the optimal action set ($ A=\left \{ a_0,a_1,..,a_{j} \right \} $) to reach the target by leveraging multimodal information gathered from the environment. The process is defined as a partially observable Markov decision process $\left \langle \mathcal{S,O,A,P,R},\gamma \right \rangle$, where $\mathcal{S}$ is the set of states, $\mathcal{O}$ is the observation space, $\mathcal{A}$ is the action space, $\mathcal{P}: \mathcal{S} \times \mathcal{A} \to \mathcal{S^{\prime}}$ is the transition probability, $ \mathcal{R}: \mathcal{S} \times \mathcal{A} \to \mathcal{R} $ is the reward function, and $\gamma$ is the discount factor. Each element is defined in detail as follows:

1) Observations: The observations include two parts as shown in Fig. 3. At timestep $t$, for visual data, it consists of a sequence of three consecutive frames $I$ captured by the camera on the bronchoscope  $\mathcal{O}^{v}_t = \{I_{t-2}, I_{t-1}, I_t\}$, and for proprioceptive data, it is a array composed of the concatenated coordinates of the bronchoscope backbone $\mathcal{O}^{p}_t = \{B_1, B_2, ..., B_{n} \}$, where $B_i = [x_i, y_i, z_i]$ denotes the position in world coordinate systerm.

2) Actions: The outer sheath and inner endoscope share the same driving mechanism. As shown in Fig. 2(b), we define discrete action elements. Specifically, at each time step, the sheath and endoscope can either move forward/backward by 3mm, or bend by 0.2 radians in $xOz$ and $yOz$ planes relative to respective coordinate systems $O_{s}xyz$ and $O_{e}xyz$. Only one of the sheath and endoscope can execute an action at the same time step. Then the action space can be defined as $\mathcal{A}=\mathcal{A}_s + \mathcal{A}_e $, where $s$ denotes the sheath and $e$ denotes the endoscope, and $\mathcal{A}_s=\left \{ s_{\it FORWARD},s_{\it BACKWARD},s_{\it LEFT},s_{\it RIGHT},s_{\it IN},s_{\it OUT} \right \}$,  $\mathcal{A}_e= \left \{ e_{\it FORWARD},e_{\it BACKWARD},e_{\it LEFT},e_{\it RIGHT},e_{\it IN},g_{\it OUT} \right \}$.

3) Rewards: By leveraging the bronchial tree, the path to the target position can be uniquely determined. The design of the reward function aims to encourage the robot to move accurately, efficiently and safely along the reference path:
$$ 
R_{t}=\omega_{1} * r_{d}+\omega_{2} * r_{a} + r_b + r_{g}, \eqno{(1)} 
$$
where
$$
r_{d}=-\left\|B_{n-1}-g_k\right\|, \eqno{(2)}
$$
$$
r_{a}=-\left(e^{\left \langle \textbf{v}_{1}, \textbf{v}_{2} \right \rangle / \pi}-1\right), \eqno{(3)}
$$
$$
r_{b}=\left\{\begin{array}{cc}
-20, & \text { if \ break } \\
0, & \text { otherwise }
\end{array}\right., \eqno{(4)}
$$
$$
r_{g}=\left\{\begin{array}{cc}
10, & \text { if \ reached } \\
0, & \text { otherwise }
\end{array}\right.  ,\eqno{(5)}
$$
where $B_{n-1}$ denotes the location of the robot's tip, $g_k$ denotes the endpoint coordinates of k-th generation airway of the reference path, $\left \langle \textbf{v}_1,\textbf{v}_2\right \rangle$ signifies the angle between the robot's orientation and the airway's orientation. The robot's orientation is defined as the vector extending from the third to the first backbone of its tip, while the airway's orientation is determined by the vector from the starting to the ending point of its centerline within the segment. $\omega_1, \omega_2$ are hyper-parameters. Additionally, to ensure safety and efficiency, we set thresholds for contact force, direction angle, and distance between the robot's tip and the target. Exceeding these thresholds results in the premature termination of the episode and a penalty $r_b$ is applied as a consequence.

\subsection{Multimodal Information Extraction and Fusion}

As previously discussed, directly acquired multimodal information differs in dimensions and values, making it unsuitable for direct input to the decision model. In this section, we delve into the representation and fusion of multimodal information.

As shown in Fig. 3, in \textbf{\textit{stage} I}, we manually operate the simulator to thoroughly explore the airways. For visual data, continuous video frames are captured, and a dynamic prediction task is designed to understand the patterns of image changes and deduce the robot's motion state. At timestep $t$, frames $I_{t-2}$, $I_{t-1}$, and $I_t$ are respectively encoded and input into the LSTM \cite{r40} model. The process can be formulated as  $r_{t}^{image} = LSTM_{\delta}(f_{\xi}(I_{t-2}), f_{\xi}(I_{t-1}), f_{\xi}(I_{t}))$, and the decoder process aims to predict the next frame: $\hat{I}_{t+1} = g_{{\xi}^{\prime}}(r_{t}^{image})$, where ${\delta}$, ${\xi}$, ${\xi}^{\prime}$ denotes the parameters of LSTM, visual encoder and decoder separately. For proprioceptive data, we use vanilla Autoencoder \cite{r42} for reconstruction task, at timestep t, the feature of proprioceptive data is $r_{t}^{proprio}=f_{\psi}(P_t)$, and the decoding process is $\hat{P_t}=g_{{\psi}^{\prime}}(r_{t}^{proprio})$, where ${\psi}$, ${{\psi}^{\prime}}$ denotes the parameters of proprioceptive autoencoder. All parameters are update by gradient descent on the reconstruction error:

$$
\xi^{\star}, \xi^{\prime \star}, \delta^{\star}=\arg \min _{\xi, \xi^{\prime}, \delta} \frac{1}{n} \sum_{i=1}^{n} \mathcal{L}\left[{I_{t+1}}^{(i)}, \hat{I}_{t+1}^{(i)} \right] , \eqno{(6)}
$$

$$
\psi^{\star}, \psi^{\prime \star}=\arg \min _{\psi, \psi^{\prime}} \frac{1}{n} \sum_{i=1}^{n} \mathcal{L}\left[{P_t}^{(i)}, \hat{P}_t^{(i)} \right] , \eqno{(7)}
$$
in which $\mathcal{L}$ is a mean squared error loss function.

In \textbf{\textit{stage} II}, we use cross-attention to fuse visual and proprioceptive information. Cross-attention enables a better capture of dynamic inter-modal relationships, which is calculated by:
\begin{align*}
& r^{state}_t = \\
& \operatorname{softmax}\left(\frac{\left(W_{Q} r^{proprio}_t\right)\left(W_{K} r^{image}_t\right)^{T}}{\sqrt{d}}\right) W_{V} r^{image}_t   , \tag{8}
\end{align*} 
where $r^{state}_t$ denotes the state feature, $W_Q$, $W_K$ and $W_V$ denote the weight matrices, and $d$ is the dimension of the vector. We continue to use the tasks mentioned above to train the attention model, ensuring that the fused vector can still individually reconstruct the original multimodal inputs. During the training process, the parameters of \textbf{\textit{stage }I} are frozen. 

   \begin{figure*}[thpb]
      \centering
      \vspace{5pt}
      \includegraphics[width=0.98\linewidth]{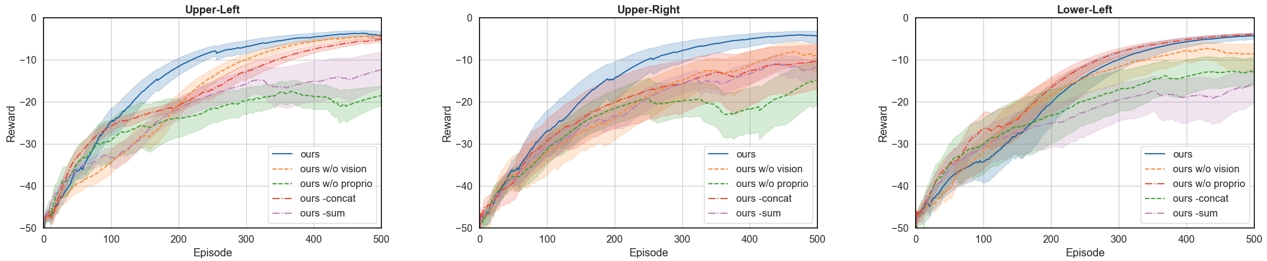}
      \caption{RL learning curves for ablation and comparison experiments: (1) BronchoCopilot, (2) BronchoCopilot without visual data, (3) BronchoCopilot without proprioceptive data, (4) BronchoCopilot using concat for fusion, (5) BronchoCopilot using sum for fusion. All curves are smoothed by exponential smoothing with a factor of r=0.95.}
      \label{figurelabel}
   \end{figure*}

   \begin{figure*}[b]
      \centering
      \includegraphics[width=0.98\linewidth]{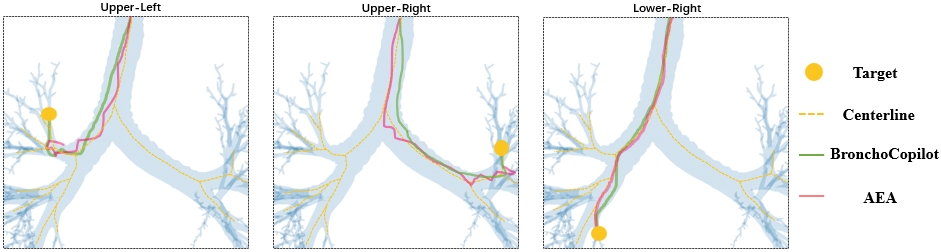}
      \caption{The target positions in the fifth-level airways were selected for the upper left, upper right, and lower left lung lobes, respectively. The yellow, green, and pink lines represent the centerlines (reference paths), BronchoCopilot, and AEA robot tip trajectories. For visualization, all trajectories represent the average of three test runs.}
      \label{figurelabel}
   \end{figure*}

\subsection{Reinforcement Learning Model}

As shown in Fig. 3, in \textbf{\textit{stage }III}, the agent receives inputs from the state representation model and updates parameters as described above. The gradient updates of RL module are also propagated to the state representation model for further fine-tuning of its parameters.

The principle of RL is to train an agent following the policy $\pi (a|s)$ which maximize the expected reward r by selected an action $a_t$ based the state $s_t$ at time step $t$. The policy $\pi ( a|s )$ is parameterized by $\theta$ and defined as $\pi_{\theta} ( a|s )$. Following the setup, $\theta$ is optimized to maximize the expected return in a policy gradient algorithm: 

$$
\theta^{*}=\mathop{\arg\max}\limits_{\theta}\left[\sum_{t} \gamma^{t} R\left(s_{t}, a_{t}\right)\right] . \eqno{(9)}
$$

We use Proximal Policy Optimization (PPO) \cite{r41} algorithm to maximize the loss function:
$$
\begin{aligned}
J_{t}^{CLIP'}\left(\theta, \theta'\right) = \mathbb{E}\left[J_{t}^{CLIP}(\theta) - c_{1} J_{t}^{VF}\left(\theta'\right) + c_{2} H\left(\pi_{\theta} \mid s_{t}\right)\right]
\end{aligned}, \eqno{(10)}
$$

where $H(\pi_{\theta}|  s_t )$ is an entropy term to encourage exploration, and $c_1$, $c_2$ are weights of loss, and $J_{t}^{VF}$ is the error term on the value estimation with discount factor $\gamma$ and target value function:
$$
J_{t}^{VF}\left(\theta^{\prime}\right)=\left(V_{\theta^{\prime}}\left(s_{t}\right)-\left(R\left(s_{t}, a_{t}\right)+\gamma V_{\theta \text { target }}\left(s_{t+1}\right)\right)\right)^{2} .\eqno{(11)}
$$
$J_{t}^{CLIP}(\theta)$ is the loss limited by a clipped ratio $\epsilon$ to stabilize the update procedure:
$$
\begin{array}{c}
J_{t}^{CLIP}(\theta)=\mathbb{E}_{t}\left[\operatorname {min} \left(r_{t}(\theta) \hat{A}_{t}(s, a), \operatorname{clip}\left(r_{t}(\theta), 1-\epsilon,\right.\right.\right. \\
\left.\left.1+\epsilon) \hat{A}_{t}(s, a)\right)\right] ,
\end{array}  , \eqno{(12)}
$$
where $r_{t} (\theta)=\pi_{\theta} (a | s)/\pi_{\theta old} (a | s)$ is the probability ratio between current policy and old policy, and $\hat{A}_t(s,a) $ is the advantage estimator calculated according to \cite{r47}.

\section{EXPERIMENT}

In this section, we perform qualitative and quantitative evaluation of our approach in the simulation environment.

\subsection{Experimental Setup}

We compare our BronchoCopilot method against prior collision-aware methods: AEA \cite{r8} and DQNN \cite{r13}. Since both methods were designed for traditional bronchoscope platform, we afford a convenience for these by providing prior information: only sheath actions are allowed before reaching the 3-th generation airways, and subsequently only endoscope actions are permitted. This adjustment more closely mirrors the typical procedural routines of clinicians \cite{r4}. In contrast, our method employs a unified action space for both the sheath and endoscope, challenging the agent to independently discern implicit operational strategies. The detailed experimental setup is as follows:

\textbf{AEA}: Given the camera image and visualized centerline, insertion or bending is determined based on the observed centerline position from the camera. We directly obtain actions without converting them into specific control signals.

\textbf{DQNN}: Given the camera image, the action space from [2] is reduced to the settings in this paper (i.e., eliminating actions numbered 2, 4, 6, 8 in the original setup, and adding a backward retreat action).

\textbf{Ablation on Multimodality}: This part involves training with only visual or proprioceptive input. The encoder is trained through the same reconstruction task and serves as the state vector, while the fusion module is removed.

\textbf{Ablation on Fusion Module}: We replace the cross-attention module with concatenation (concat) and summation (sum) individually while keeping other settings consistent.

\begin{table*}[t]
\vspace{5pt}
\caption{\textbf{Overall performance comparison}. We evaluate 7 methods, including AEA, DQNN, BronchoCopilot without proprioception or vision, and using concat or sum for fusion in 3 tasks.}
\huge
\renewcommand{\arraystretch}{1.3}
\setlength{\heavyrulewidth}{1.3pt} 
\setlength{\lightrulewidth}{1.3pt}   
\resizebox{\linewidth}{!}{
\begin{threeparttable} 
\begin{tabular}{@{} l *{12}{c} @{}}
\toprule
& \multicolumn{4}{c}{\textbf{Upper-Left}} & \multicolumn{4}{c}{\textbf{Upper-Right}} & \multicolumn{4}{c}{\textbf{Lower-Left}} \\
\cmidrule(lr){2-5} \cmidrule(lr){6-9} \cmidrule(lr){10-13}
& SR(\%)↑ & NA↓ & TL↓ & F\_M/A(10\textasciicircum{}(-1)N)↓ & SR↑ & NA↓ & TL↓ & F\_M/A(10\textasciicircum{}(-1)N)↓ & SR↑ & NA↓ & TL↓ & F\_M/A(10\textasciicircum{}(-1)N)↓ \\
\midrule
\textbf{AEA} & 54.3(±6.4) & - & - & - & 76.2(±3.3) & 322.2(±12.5) & 1.24(±0.15) & 2.78/0.24 & 89.8(±1.8) & \textbf{251.4(±19.7)} & \textbf{0.94(±0.15)} & \textbf{3.34/0.22} \\
\textbf{DQNN} & 0 & - & - & - & 0 & - & - & - & 23.8(±7.7) & - & - & - \\
\textbf{ours w/o P} & 43.6(±6.7) & - & - & - & 57.7(±4.9) & - & - & - & \textbf{92.6(±1.7)} & 255.0(±30.7) & 0.98(±0.16) & 4.34/0.88 \\
\textbf{ours w/o V} & 91.2(±3.8) & 272.2(±18.5) & 1.39(±0.17) & \textbf{1.88/0.43} & 73.3(±5.8) & 332.2(±19.6) & 1.15(±0.18) & 11.22/1.44 & 86.7(±4.2) & 298.6(±31.0) & 1.03(±0.17) & 12.10/1.57 \\
\textbf{ours-concat} & 91.5(±2.1) & 255(±24.3) & 0.99(±0.08) & 3.44/0.25 & 82.7(±1.1) & 344.4(±25.2) & 1.03(±0.07) & 1.55/0.38 & 63.3(±2.6) & - & - & - \\
\textbf{ours-sum} & 83.4(±4.2) & 352.2(±23.1) & 1.32(±0.09) & 9.87/0.74 & 85.6(±2.8) & 328.1(±34.6) & 1.07(±0.11) & 2.94/0.77 & 28.6(±8.2) & - & - & - \\
\textbf{ours} & \textbf{97.1(±1.2)} & \textbf{223.2(±12.7)} & \textbf{0.96(±0.04)} & 2.56/0.38 & \textbf{95.5(±0.8)} & \textbf{289.3(±8.4)} & \textbf{0.95(±0.06)} & \textbf{1.67/0.25} & 91.3(±2.4) & 269.6(±23.8) & 1.01(±0.13) & 5.09/0.97 \\
\bottomrule
\end{tabular}
\begin{tablenotes}
        \Huge
        \centering
        \item `-' indicates that the metric is not calculated due to a low success rate.
      \end{tablenotes}
  \end{threeparttable}
}
\end{table*}

Based on the reconstructed 3D model, we selected three targets in the 5-th generation airways of the upper left, lower left, and upper right lung lobes for training. During the model training phase, we evaluated the agent's accumulated reward and learning efficiency, which are depicted in the reward curve shown in Fig. 4. In the model evaluation phase, we conducted 80 insertion procedures, with the targets' locations and insertion paths illustrated in Fig. 5. The metrics for evaluation include: (i) Success Rate (${\it SR}$), defined when the robot's tip is within 7mm of the target. This threshold corresponds to the length of the standard biopsy tool attached to the bronchoscope, as shown in Fig. 1(a). A task is considered a failure if, after 500 actions, the robot has not reached the target or has triggered the exit criteria described in Section III. (ii) Number of Actions (${\it NA}$), representing the total actions taken to reach the target. (iii) Trajectory Length (${\it TL}$), recording the distance the robot's tip travels. To facilitate comparison across different experiments, all lengths are normalized by the reference path length to the target. (iv) The Maximum and Average contact Forces (${\it F_{M/A}}$), while contact is unavoidable, minimizing force is preferable. 

   \begin{figure}[]
      \centering
      \includegraphics[scale=0.4]{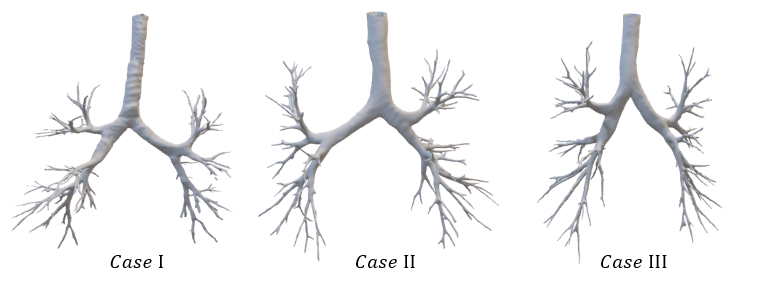}
      \caption{Different airway models for transfer training.}
      \label{figurelabel}
   \end{figure}

\begin{table}[h]
\renewcommand{\arraystretch}{0.85} 
\centering
\caption{Mean value of Success Rate for different airway structures and its generations.}
\begin{tabular}{cccccc}
\toprule
            & \textbf{2-th} & \textbf{3-th} & \textbf{4-th} & \textbf{5-th} & \textbf{6-th} \\ 
\midrule
\textbf{Case I}   & 100.0\%     & 96.6\%      & 91.2\%     & 83.3\%    & 59.8\%     \\
\textbf{Case II}  & 100.0\%    & 92.4\%       & 83.5\%      & 79.3\%     & 43.7\%      \\
\textbf{Case III} & 100.0\%     & 94.2\%       & 88.7\%      & 85.4\%     & 76.6\%      \\ 
\bottomrule
\end{tabular}
\end{table}

For \textbf{Transfer Experiment}, we fixed the parameters of \textit{stage I \& II} and exclusively trained the RL module. We randomly selected CT images of three cases from the EXACR'09 dataset \cite{r46} and performed the segmentation process as described in Section III. A. The anatomical structures of the cases are depicted in Fig. 6. We randomly selected three targets on the centerlines of airways at different levels for training. Subsequently, we conducted 80 evaluation trials to determine the success rate, with the findings presented in Table II.

\subsection{Implementation Details}

For all networks involved in our experiments, we employed Kaiming initialization \cite{r44} for the weight matrices. The camera images were captured at a resolution of 512x512, and we utilized ResNet34 \cite{r45} as the visual encoder alongside a Multilayer Perceptron (MLP) for the proprioceptive encoder. The state representation model outputs a vector of dimension 64 (or dimension 128 when concatenated). In the reinforcement learning (RL) model, we implemented two MLPs for the actor and critic networks, respectively. Each network consists of five layers with 256 nodes and employs Tanh as the activation function. Our experiments were conducted using the PyTorch framework on a workstation equipped with an Intel i7-13700KF CPU and an NVIDIA RTX 4070 GPU. During \textit{stages I \& II}, we trained for 50 epochs on a dataset comprising approximately 30,000 images. For \textit{stage III}, the training extended over 500 episodes, with each episode capped at a maximum of 1000 steps. The average training duration was 2.2 hours, with the model typically reaching convergence between the 120th and 190th episodes.

\section{RESULT AND DISCUSSION}

\textbf{The BronchoCopilot outperforms previous image-guided and RL-based bronchoscopy agents, with multimodal information significantly contributes to the improved performance.} As shown in Table I and Fig. 5, in all three tasks, BronchoCopilot performs excellently. Compared to AEA and DQNN, our approach demonstrates nearly the best performance across four metrics in the upper lobe task. While AEA shows commendable results in the lower lobe task, where the centerline remains visible throughout the insertion without major turns, it struggles with the pre-bend strategy essential for accessing the upper lobe. DQNN, with its overly simplistic network design, fails to adequately capture image change patterns, leading to convergence issues during our training sessions.  

We observed severe algorithmic failures when the robot's tip is very close to the airway wall. Conversely, BronchoCopilot leverages proprioceptive data to bolster decision-making processes, achieving a success rate exceeding 90\% for insertions into the 5-th generation airways. The LSTM-based visual module plays a crucial role in detecting shifts in imagery,  which minimizes oscillations during insertion and contributes to a reduction in both ${\it NA}$ and ${\it TL}$. Our method exhibits minimal variance across most evaluation metrics, indicating that BronchoCopilot can execute highly consistent actions. This consistency is vital for minimizing the risk of unforeseen events during surgeries. In tasks involving the lower lobe, where the pathway is smoother and requires less bending, our method is slightly inferior to more intuitive approach (AEA) due to considering more factors.

\textbf{The fusion approach of visual and proprioceptive information outperforms traditional fusion methods.} Through comparison tests between concatenation and summation methods, it's evident that BronchoCopilot's incorporation of cross-attention mechanisms significantly enhances performance. This cross-attention functionality enables the model to autonomously discern the interplay between modalities and dynamically modulate the weight assigned to each, ensuring an optimal blend of information for decision-making. The efficacy of this approach is clearly validated in the results presented in Table I, showcasing the robust advantage of cross-attention in multimodal information fusion.

\textbf{Our method is capable of rapid end-to-end training transfer across diverse surgical cases.} By freezing the parameters of the state representation model and focusing on training the decision model in an end-to-end fashion, we've shown that BronchoCopilot consistently achieves high success rates across diverse anatomical structures. Furthermore, the average training duration for tasks targeting the 5-th generation airways stands at merely 0.68 hours. This efficiency in end-to-end transfer training underscores our method's swift adaptability to different bronchoscopy cases, emphasizing its significant practical utility in the field.

\section{CONCLUSION AND FUTURE WORK}

In this paper, we introduced BronchoCopilot, a multimodal reinforcement learning algorithm and training framework designed for the autonomous robotic bronchoscopy. Leveraging the synergistic potential of visual and proprioceptive information, BronchoCopilot represents a significant advancement in the manipulation of dual-segment flexible bronchoscopy robot, particularly in more complex airway environments. Our staged training approach not only simplifies the complexity associated with training multimodal RL models but also enables rapid adaptation to diverse surgical scenarios through end-to-end transfer learning. As we look to the future, the translation of BronchoCopilot to real-world clinical settings and its validation on physical robotic platforms stand as the next frontier in our research. 



\bibliography{ref} 

\end{document}